\documentclass{amam}                
\usepackage[utf8]{inputenc}         
\usepackage{textcomp}               
\usepackage{graphicx}               
\usepackage{url}                    

\usepackage{epsfig} 
\usepackage{times} 
\usepackage{amsmath} 
\usepackage{amssymb}  
\usepackage{nicefrac}
\usepackage{tikz}
\usepackage{pgfplots}
\pgfplotsset{compat=1.18} 

\usepackage{comment}    

\usepackage{bm}        
\usepackage{graphicx}  
\usepackage{bbold}     

\usepackage{enumerate}
\usepackage[backend=biber, url=false, doi=false]{biblatex}
\newcommand{\kl}{k_\mathrm{l}}

\newcommand{\T}{^{\mathop{\mathrm{T}}}}

\numberwithin{thm}{subsection}

\title{How Leg Stiffness Affects Energy Economy in Hopping}

\author{Iskandar Khemakhem, Dominik Tschemernjak,
        Maximilian Raff, C. David Remy\\
Institute for Adaptive Mechanical Systems, University of Stuttgart, Germany\\
{\it khemakhem@iams.uni-stuttgart.de} \\
}

\begin{document}
\begin{titlepage}
\centering


This preprint has been published as:\\
\textit{How Leg Stiffness Affects Energy Economy in Hopping}\\[0.3cm]

Iskandar Khemakhem, Dominik Tschemernjak, Maximilian Raff, and C. David Remy\\[0.3cm]

Presented at: \textit{12th International Symposium on Adaptive Motion of Animals and Machines (AMAM 2025)}\\
Darmstadt, Germany, July 7--11, 2025\\[0.3cm]

DOI: \url{https://doi.org/10.26083/tuprints-00030955}\\[0.5cm]

Please cite the published version:\\
Khemakhem, I., Tschemernjak, D., Raff, M., \& Remy, C. D. (2025). How leg stiffness affects energy economy in hopping. In \textit{12th International Symposium on Adaptive Motion of Animals and Machines (AMAM 2025)}, Darmstadt, Germany. TUprints.

\vfill

\end{titlepage}

\maketitle

\section{Motivation}
Humans and animals demonstrate impressive agility, versatility, and efficiency in legged locomotion. By employing a wide range of gaits, they easily adapt to varying speeds and terrains. 
In each gait, biological systems cleverly harness their bodies' mechanical dynamics to minimize energy usage and enhance performance. 
One key aspect in energetically economic locomotion is the strategic usage of leg compliance, leveraging elastic structures such as muscles, tendons and ligaments to store and release energy \cite{dickinson2000how}.

In the robotics community, we aim to achieve similar performance with legged robots.
However, creating robotic legs that perform consistently across diverse operating conditions, particularly at varying average forward speeds remains a challenge.
It remains uncertain whether achieving this requires adapting the stiffness of elastic elements, or if similar performance can be achieved by adjusting motion and actuation while keeping the stiffness fixed.

To explore this question, we model a monopedal robot featuring a spring operating in parallel to the leg's actuator and explore the influence of the leg stiffness on the energy economy when performing periodic hopping motions. 
To this end, we formulate an optimal control problem to find periodic motions and actuator inputs that minimize the energetic cost of transport for a given forward speed and leg stiffness. 
We perform a parameter study by solving the optimal control problem using direct collocation over a grid of average forward speeds and leg stiffness values. 

This work gives insights into efficient design of robotic legs and their adaptation to varying conditions. It also contributes to the broader idea of using trajectory optimization to co-design a robot's morphology alongside its controller. 

\section{Methods}
\subsection{Parameterized Trajectory Optimization}\label{sec:Parameterized Trajectory Optimization}
\begin{figure}
    \centering
    \includegraphics{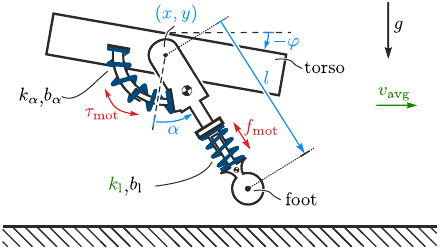}
    \caption{A monoped with parallel elastic actuation in the leg and hip joint.}
    \label{fig:monoped}
\end{figure}

We consider the planar monopedal robot illustrated in Figure~\ref{fig:monoped}, based on the model from \cite{raff2022generating}. 
The robot has five degrees of freedom, described by the generalized coordinates~$\bm{q}\T = [x~y~\varphi~\alpha~l]$, representing the torso's pose, the hip angle relative to the torso, and the leg's overall length. 
Actuation is achieved through a torque~$\tau_\mathrm{mot}$ at the hip and a translational force~$f_\mathrm{mot}$ at the leg joint. 
Both actuators are arranged in parallel with springs and dampers, defined by stiffness~$k_\alpha$ and~$\kl$, and damping~$b_\alpha$ and~$b_\mathrm{l}$, with damping proportional to stiffness.
Following \cite{hof1996scaling}, model parameters are normalized using the total body mass~$m$, natural leg length~$l_\circ$, and gravity~$g$. 
The hopping gait consists of two phases: \textit{Stance}, where the foot is in contact with the ground, and \textit{Flight}, where the hopper is airborne. Together, these phases form a hybrid dynamical system, with flow maps connected via an event set and reset map, as described in~\cite{raff2022connecting}.

We formulate an optimization problem to determine the monoped's optimal motion, minimizing the cost of transport (CoT), defined as the squared input normalized by stride distance. 
The optimization is performed over a single stride.
Periodic motion is enforced by requiring all initial states, except for the forward position to repeat at the stride's end.
A stride begins at touch-down, marking the start of the stance phase. 
During stance, the foot is constrained to pure rolling motion.
Once ground contact is lost, the robot enters the flight phase. 
The stride ends when the foot regains contact with the ground, after applying the reset map to the generalized velocities resulting from the impact. 
We specify an operating point by fixing the average forward speed, $v_\text{avg}$.

We define two parameterized optimization problems: 
$\mathcal{P}_1(\textcolor[rgb]{0.09,0.51,0}{v_\text{avg}})$ determines the optimal motion, actuator inputs, and leg stiffness $\kl$ for a given forward speed $v_\text{avg}$,
whereas $\mathcal{P}_2(\textcolor[rgb]{0.09,0.51,0}{v_\text{avg}},\,\textcolor[rgb]{0.09,0.51,0}{\kl})$ determines the optimal motion when both $v_\text{avg}$ and $\kl$ are fixed.

\subsection{Numerical Implementation}
Both optimization problems are discretized via direct collocation with a Hermite–Simpson scheme, forming nonlinear programs (NLPs) as described in~\cite{kelly2017introduction}. The stance and flight time domains are each divided into~$N=30$ equidistant segments, while the total duration of each phase is treated as a decision variable in the optimization. Dynamics are enforced as constraints at the midpoints of each segment, with the additional constraints mentioned in Section~\ref{sec:Parameterized Trajectory Optimization} imposed at the start and end of each phase.

The dynamics in each phase are extended by including the segment duration and leg stiffness $\kl$ as additional states, with zero derivatives. Additionally, the actuator inputs are treated as states, and their derivatives are introduced as inputs to the extended dynamics. The extended states are assumed to be piecewise cubic, while the inputs of the extended dynamics are modeled as piecewise linear. This formulation ensures continuously differentiable, piecewise quadratic inputs and avoids a highly coupled and dense Jacobian for the constraints \cite{kelly2017transcription}.

We follow heuristics to generate suitable initial conditions, which are necessary to avoid landing in unwanted local minima, as the resulting NLPs are nonlinear, non-convex, and high-dimensional. We start by solving $\mathcal{P}_1$ for a given $v_\text{avg}$ to obtain an optimal motion and the associated optimal leg stiffness. The resulting solution is then used to explore neighboring points on a fine grid of leg stiffness values ($\kl \in [1,\,13]\,mg/l_\circ$, $\Delta \kl = 0.1\,mg/l_\circ$) and forward velocities ($v_\text{avg} \in [0.05,\,1.45]\,\sqrt{l_\circ g}$, $\Delta v_\text{avg} = 0.01\,\sqrt{l_\circ g}$) 
by solving $\mathcal{P}_2$. We iteratively extend this process to explore the parameter space, generating a library of optimal gaits.

\section{Results}
\begin{figure}[h]
    \centering
    \includegraphics{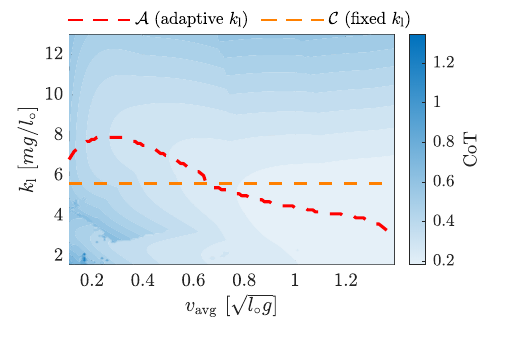}
    \caption{Top view of the map of optimal gaits in the $(v_\mathrm{avg}, \kl)$-space, colored by cost of transport. The slices \textcolor[rgb]{1,0,0}{$\mathcal{A}$} (adaptive stiffness) and \textcolor[rgb]{1,0.5,0}{$\mathcal{C}$} (fixed stiffness) are highlighted.}
    \label{fig:k_over_v}
\end{figure}

Figure~\ref{fig:k_over_v} depicts the evolution of the CoT across the parameter space. For further analysis, we examine two slices of this space:~$\mathcal{A}$ traces the gaits with the lowest CoT at each forward speed, and $\mathcal{C}$ uses a constant stiffness equal to the mean of the optimal stiffness values in $\mathcal{A}$.
Figure~\ref{fig:k_over_v} reflects the long established concept that, at lower speeds a stiffer leg enables more efficient locomotion, whereas compliance gains significance at higher speeds.
Additionally, the region surrounding $\mathcal{A}$ is relatively flat, indicating minimal variation in cost when selecting a slightly different stiffness value.

\begin{figure}
    \centering
    \includegraphics{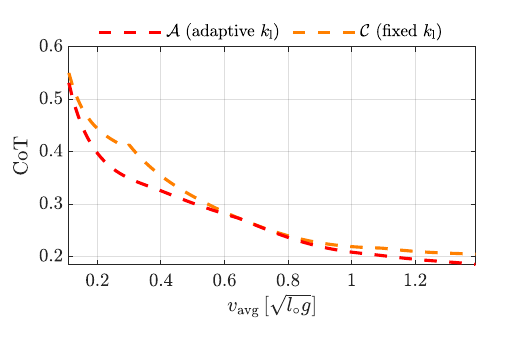}
    \caption{Comparison of the cost of transport between the slices \textcolor[rgb]{1,0,0}{$\mathcal{A}$} and \textcolor[rgb]{1,0.5,0}{$\mathcal{C}$} across various average forward speeds.}
    \label{fig:cot_over_v}
\end{figure}

Figure~\ref{fig:cot_over_v} compares the CoT for $\mathcal{A}$ and $\mathcal{C}$ across varying forward speeds. 
Since for the slice $\mathcal{A}$, the stiffness is optimized at each speed, it results in a lower CoT compared to $\mathcal{C}$.
Still, the two curves remain closely aligned, with the most significant increase in CoT observed at $v_\mathrm{avg} \approx 0.3 \sqrt{l_\circ g}$, where the CoT rises by $20\%$. 
Averaged across all speeds, the CoT increases by $6\%$ when using a constant leg stiffness (slice~$\mathcal{C}$). Notably, the curves nearly overlap in the speed range $v_\mathrm{avg} \in [0.6,\,0.8] \sqrt{l_\circ g}$, indicating that adapting the leg stiffness has little impact on the cost in this region.

\section{Conclusions}
While adapting the leg stiffness to the robot's speed may provide energy savings of up to $20\%$ at certain speeds, the overall improvement in energy economy for a monoped with adaptive leg stiffness is modest compared to a well-chosen fixed stiffness. 
Although this improvement provides a compelling enough argument for implementing variable stiffness in some cases, the practical challenges of doing so—including additional energy consumption, higher costs, and the need for sophisticated control strategies— might ultimately outweigh the gains \cite{hurst2008role}.
This work concludes, that a carefully chosen constant stiffness achieves similar results in most scenarios.

The non-convexity of the optimization problems limits our approach to ensuring only local minima.
Future work will further explore this parameter space, also including the hip stiffness~$k_\alpha$.

\section{Acknowledgments}

This work was funded by the Deutsche Forschungsgemeinschaft (DFG) – Projects 501862165 and 533240481. It was further supported through the International Max Planck Research School
for Intelligent Systems (IMPRS-IS) for Iskandar Khemakhem and Maximilian Raff.

\printbibliography
%
%

\end{document}